\documentclass[letterpaper]{article} 
\usepackage{aaai25}  
\usepackage{times}  
\usepackage{helvet}  
\usepackage{courier}  
\usepackage[hyphens]{url}  
\usepackage{graphicx} 
\usepackage{natbib}  
\usepackage{caption} 
\usepackage{algorithm}
\usepackage{algorithmic}
\usepackage{multirow}
\usepackage{cuted}
\usepackage{enumitem}
\usepackage{xcolor}

\usepackage{newfloat}
\usepackage{listings}
\usepackage{booktabs}
\DeclareCaptionStyle{ruled}{labelfont=normalfont,labelsep=colon,strut=off} 
\floatstyle{ruled}
\newfloat{listing}{tb}{lst}{}
\floatname{listing}{Listing}
\usepackage{amsmath,amsfonts,amssymb}
\DeclareMathAlphabet{\mathbbold}{U}{bbold}{m}{n}

\usepackage{cleveref}

\nocopyright

\title{\raisebox{-0.27\height}{\includegraphics[height=1.7em]{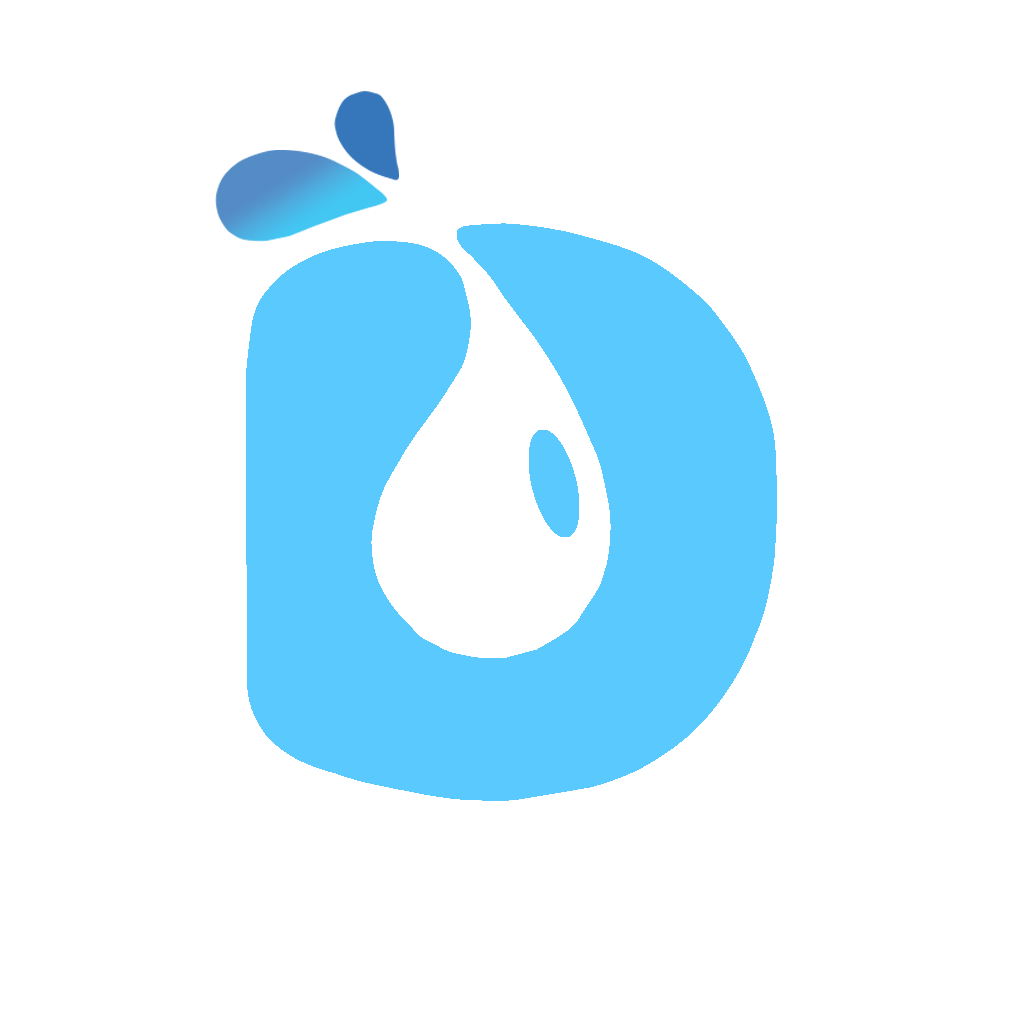}}\hspace{-0.3em}eRain{\color{cyan}GS}: Gaussian Splatting for Enhanced Scene Reconstruction in Rainy Environments}
\author{
    Shuhong Liu\textsuperscript{\rm 1},
    Xiang Chen\textsuperscript{\rm 2}, Hongming Chen\textsuperscript{\rm 3}, Quanfeng Xu\textsuperscript{\rm 4,5}, Mingrui Li\textsuperscript{\rm 6}\thanks{Corresponding Author}
}
\affiliations{
    \textsuperscript{\rm 1}The University of Tokyo, \textsuperscript{\rm 2}Nanjing University of Science and Technology\\
    \textsuperscript{\rm 3}Dalian Maritime University, \textsuperscript{\rm 4}Shanghai Astronomical Observatory\\
    \textsuperscript{\rm 5}University of Chinese Academy of Sciences, 
    \textsuperscript{\rm 6}Dalian University of Technology
}

\usepackage{bibentry}

\begin{document}

\maketitle

\begin{strip}
\vspace{-11mm}
\centering
\noindent
\includegraphics[width=\linewidth]{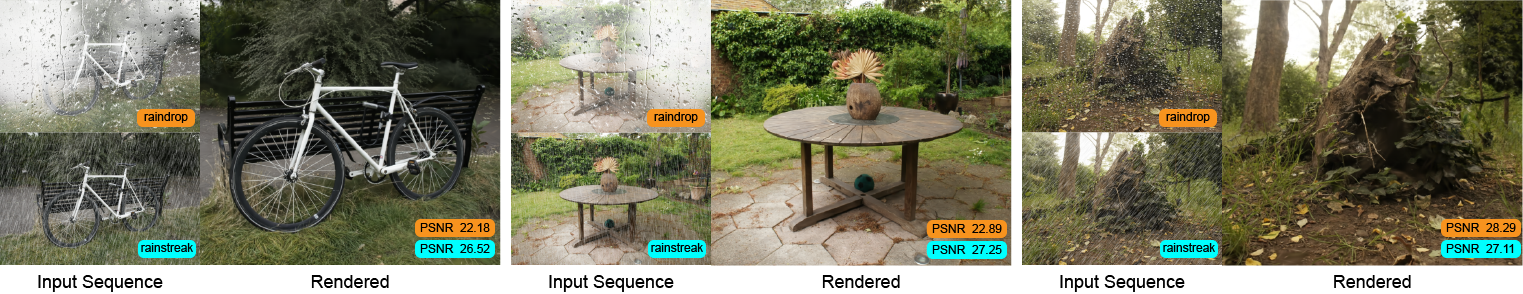}
\captionof{figure}{DeRainGS reconstructs clean scenes from input rainy images. The visualization displays three different scenes along with the reconstruction outcomes of DeRainGS on our HydroViews dataset.}
\label{fig:abstract-figure}
\end{strip}

\begin{abstract}
Reconstruction under adverse rainy conditions poses significant challenges due to reduced visibility and the distortion of visual perception. These conditions can severely impair the quality of geometric maps, which is essential for applications ranging from autonomous planning to environmental monitoring. In response to these challenges, this study introduces the novel task of 3D Reconstruction in Rainy Environments (3DRRE), specifically designed to address the complexities of reconstructing 3D scenes under rainy conditions. To benchmark this task, we construct the HydroViews dataset that comprises a diverse collection of both synthesized and real-world scene images characterized by various intensities of rain streaks and raindrops. Furthermore, we propose DeRainGS, the first 3DGS method tailored for reconstruction in adverse rainy environments. Extensive experiments across a wide range of rain scenarios demonstrate that our method delivers state-of-the-art performance, remarkably outperforming existing occlusion-free methods by a large margin. The project page is available at~{\color{cyan}\url{https://deraings.github.io}}.
\end{abstract}

%

\section{Introduction}

Images and videos captured in rainy conditions often exhibit significant quality degradation due to reduced visibility and distortions such as rain streaks and waterdrops. While such conditions pose challenges across various fields, they are particularly problematic for 3D reconstruction tasks, which rely on clear and accurate visual inputs to effectively model environments. In practical applications like autonomous vehicles \cite{hnewa2020object}, aerial surveying \cite{chang2024uav}, and outdoor robotics \cite{fumagalli2021fast}, rain can severely distort the geometry, depth, and textures that are critical for accurate mapping, leading to blurred details and spatial distortions.

Remarkable strides in 3D scene reconstruction, notably through Neural Radiance Field (NeRF) \cite{mildenhall2021nerf} and 3D Gaussian Splatting (3DGS) \cite{kerbl20233d}, have substantially enhanced the quality and efficiency of 3D scene representation and rendering. NeRF employs a neural implicit representation using Multi-layer Perceptron (MLP), providing a continuous volumetric scene representation that enables high-quality novel view synthesis (NVS). In contrast, 3DGS utilizes an explicit representation with spherical harmonics (SH) to model scene geometry and appearance, delivering high-fidelity reconstruction with fine-grained details while achieving exceptional rendering speed through efficient Gaussian rasterization.

Building on these advancements, recent studies \cite{martin2021nerf, chen2022hallucinated, ren2024nerf, kulhanek2024wildgaussians, xu2024wild, wang2024we} have expanded to facilitate reconstruction in unstructured, ``in-the-wild" environments, tackling challenges posed by transient or dynamic objects and adapting to versatile styles and appearances. However, these approaches have only been demonstrated to be effective in mitigating typical transient and dynamic occluders. When it comes to addressing adverse weather conditions such as rain, these methods encounter significant limitations. Rain introduces unique visual artifacts that are not merely transient distractors but continuous disturbances that affect visibility throughout the entire scene. Unlike typical occluders, which can be effectively segmented and removed \cite{yang2023cross, zhang2024gaussian}, the pervasive impact of rain on visibility and the complex interplay of light and water significantly challenge existing reconstruction techniques.

On the other hand, current image deraining approaches, which predominantly focus on single image deraining \cite{wang2019spatial, wei2021deraincyclegan, chen2022unpaired, chen2024bidirectional} or video deraining for static scenes \cite{wei2017should, jiang2018fastderain, cho2020single, yue2021semi}, face certain limitations in the context of 3D reconstruction. Despite showing efficacy in mitigating rain effects, these methods struggle to fully eliminate rain-induced distortions due to the irregular, complex nature of real-world rain streaks and the ill-posed nature of the problem. Crucially, these methods focus primarily on pixel-level manipulations in 2D images, neglecting the critical aspect of geometry accuracy that is heavily affected by rain. Naively integrating a derainer to preprocess scene images before reconstruction yields suboptimal results, as shown in \Cref{fig:derainer-artifact}.

To overcome these challenges, we introduce a new task: \textbf{ 3D Reconstruction in Rainy Environments (3DRRE)}. This novel formulation aims at reconstructing 3D scenes under adverse rainy conditions. To benchmark this task, we have developed the HydroViews dataset, which includes both synthesized and real-world scene images featuring a variety of rain streaks and raindrops. Building upon this dataset, we propose DeRainGS, a novel reconstruction pipeline tailored for the 3DRRE problems. DeRainGS integrates unique rainy image enhancement and occlusion masking modules to explicitly tackle the impact of rain on visual features and geometric consistency.

To summarize, our key contributions include:
\begin{itemize}
    \item We establish a new task in 3D reconstruction, specifically aiming at constructing clean scenes from rainy image input. This task is supported by our newly constructed HydroViews dataset, featuring both synthesized and real-world scenes to cover common practical scenarios.
    \item We propose DeRainGS, the first 3D reconstruction method utilizing 3DGS for high-quality reconstruction under adverse rainy conditions.
    \item DerainGS effectively incorporates novel rainy image enhancement and occlusion masking modules into a unified pipeline, collectively contributing to reconstructing clean scenes from rainy environments.
\end{itemize}

\section{Related Work}

\subsection{Image Deraining}
In recent years, image deraining has become a crucial preprocessing step in high-level vision tasks. It can be further categorized into single image deraining and video deraining~\cite{chen2023towards}. Video deraining utilizes temporal redundancy and rain dynamics, whereas single image deraining primarily relies on spatial information from neighboring pixels and the visual characteristics of rain and background scenes. Deep learning-based methods have become predominant in image deraining due to their powerful representational capacity compared to conventional prior-based methods. Researchers can opt for deep models based on convolutional neural networks (CNNs)~\cite{ren2019progressive,yang2019joint} or Transformer architectures~\cite{chen2023learning,chen2023sparse} to directly estimate a clear image from a rainy one. 

Compared to CNNs, Transformers have an advantage in image deraining, especially when dealing with long spatial rain effects. Transformers can effectively capture and model the dependencies over longer distances in the image, allowing them to better understand and remove complex rain patterns that extend across larger spatial areas. Xiao et al.~\cite{xiao2022image} introduce an image deraining Transformer that incorporates spatial-based and window-based self-attention modules. Recently, Chen et al.~\cite{chen2024bidirectional} are the first to explore integrating implicit neural representations into Transformers to enhance the robustness of image deraining. However, these methods can only handle specific forms of rain degradation. In real-world scenarios, rain is diverse and uncertain in terms of density, shape, location, and size. Therefore, improving the generalization capability of deraining models across different types of rain is crucial.

\subsection{3D Reconstruction in the Wild}

Rain, particularly waterdrops that adhere to camera lenses, can be treated as semi-transparent occluders to the views. Other occlusions, such as moving pedestrians and vehicles, are frequently presented in real-world image collections. Regardless of varying illumination across frames, a primary concern arises from these dynamic obstacles which can introduce substantial artifacts in reconstructed scenes. NeRF-W \cite{martin2021nerf} tackled transient occlusions with a dual NeRF setup and aleatoric uncertainty. Ha-NeRF \cite{chen2022hallucinated} uses learned priors for better occlusion reconstruction, while CR-NeRF \cite{yang2023cross} aligns features across views for improved occlusion removal. In 3DGS, Wild-GS \cite{xu2024wild} and GS-W \cite{zhang2024gaussian} use U-Net \cite{ronneberger2015u} segmentation masks for occlusion supervision. SpotlessSplats \cite{sabour2024spotlesssplats} employs an MLP classifier for dynamic distractors, and WildGaussians \cite{kulhanek2024wildgaussians} uses DINO \cite{ronneberger2015u} features to mask transient objects. While these in-the-wild methods effectively address floaters from transient obstacles, they falter with the widespread distortion and coverage caused by rain which uniquely affects visibility and scene integrity. To address this challenge, we incorporate a novel image enhancement network that is essential for mitigating rain-induced disturbances.

Two closely related works to ours, Occlusion-free NeRF \cite{zhu2023occlusion} and DerainNeRF \cite{li2024derainnerf}, attempt to mitigate rain effects in 3D reconstruction. Occlusion-free NeRF \cite{zhu2023occlusion} employed bidirectional volume rendering to distinguish foreground obstacles from the camera views. DerainNeRF \cite{li2024derainnerf} trivially combines a deraining AttGAN network \cite{qian2018attentive} with NeRF, removing the waterdrops using the hardcoded masks. However, both approaches struggle to handle dense rain streaks commonly seen in adverse weather conditions. Moreover, these methods have only been validated in a sparse-view configuration with limited camera angles, where foreground and background can be clearly separated. Such scenarios are impractical for real-world applications.


\section{Dataset Construction}
\begin{figure}[h]
\centering
\includegraphics[width=0.45\textwidth]{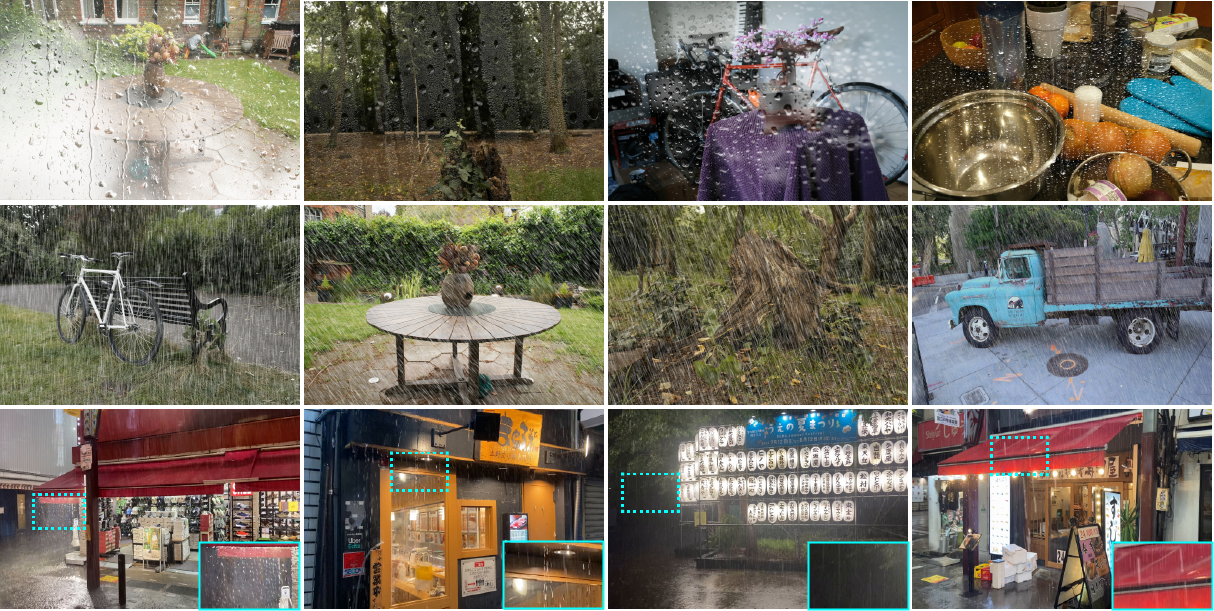}
\caption{The visualization of our HydroViews dataset. From top to bottom, it includes synthesized raindrops, rain streaks, and real-world scenes.}
\label{fig:dataset}
\end{figure}

We observe that real-world rain effects exhibit a wide range of appearance characteristics, such as variations in rain shape, length, and direction. Existing rain datasets~\cite{chen2023towards} typically create rainy images using Photoshop software to randomly add rain streaks onto clear backgrounds. In fact, synthesizing rain in 2D images and 3D scenes involves distinct processes and challenges, with a significant difference being the continuity across multiple camera views in 3D scenes. In this study, we propose a new pipeline for synthesizing rainy images tailored for 3D scenes that include both rain streaks and raindrops. As depicted in \Cref{fig:dataset}, we synthesize data derived from real-world scenes captured by MipNeRF-360~\cite{barron2022mip} and Tanks-and-Temples \cite{Knapitsch2017}, along with real-world rainy environments that we have independently collected.

\subsection{Rain Streak Generation}
We synthesize rain streaks using motion blur, which involves creating streaks that mimic the natural motion of falling rain. This technique leverages the motion blur effect to simulate the linear, directional movement of rain, capturing the essence of rain streaks by considering their repeatability and directionality. Mathematically, this can be represented as:
\begin{equation}
\mathbf{S}=\mathbf{K}(l, \theta, w) \otimes \mathbf{N}(n),
\end{equation}

\noindent where $\mathbf{N}$ represents the rain streak layers generated by random noise. We utilize uniform random numbers and thresholds to control the noise level $n$. The length and angle of the motion blur kernel $\mathbf{K} \in \mathbb{R}^{p \times p}$ are denoted by $l$ and $\theta$, respectively. Additionally, we incorporate a rotated diagonal kernel with Gaussian blur to achieve the desired rain thickness $w$. The values for noise quantity $n$, rain length $l$, rain angle $\theta$, and rain thickness $w$ are sampled from the ranges $[100,300]$, $[20,40]$, $[40^{\circ},120^{\circ}]$, and $[3,7]$, respectively. The symbol $\otimes$ denotes the spatial convolution operator. To ensure the continuity of rain synthesis in 3D scenes using the motion blur method, we keep these parameters within fixed ranges in the same 3D scene.

\subsection{Raindrop Generation}
As an atmospheric process, raindrops are individual spherical water droplets, while rain streaks are elongated trails formed by the motion blur of falling raindrops. Here, we use the open-source 3D graphics engine Blender~\footnote{https://blendermarket.com/products/rain-generator} to simulate continuous raindrops in realistic scenes, utilizing its fluid motion model and rain generator plugin to render raindrops with accurate transparency. We model the blurred or occluded effects $D$ of raindrops in dispersed, small-sized, locally coherent regions and combine these with background images $B$ to create the degraded raindrop images ${R}_{d}$. Mathematically, the synthesis process can be expressed as:
\begin{equation}
{R}_{d}=(1-M_d) \odot B+D,
\end{equation}
where $\odot$ means element-wise multiplication, $D$ denotes the occlusion or blurring effects of raindrops, and $B$ is a binary mask. When $M_d(x)=1$, the pixel $x$ in the mask is part of the raindrop region; otherwise, it belongs to the background.

\begin{figure*}[ht]
\centering
\includegraphics[width=\textwidth]{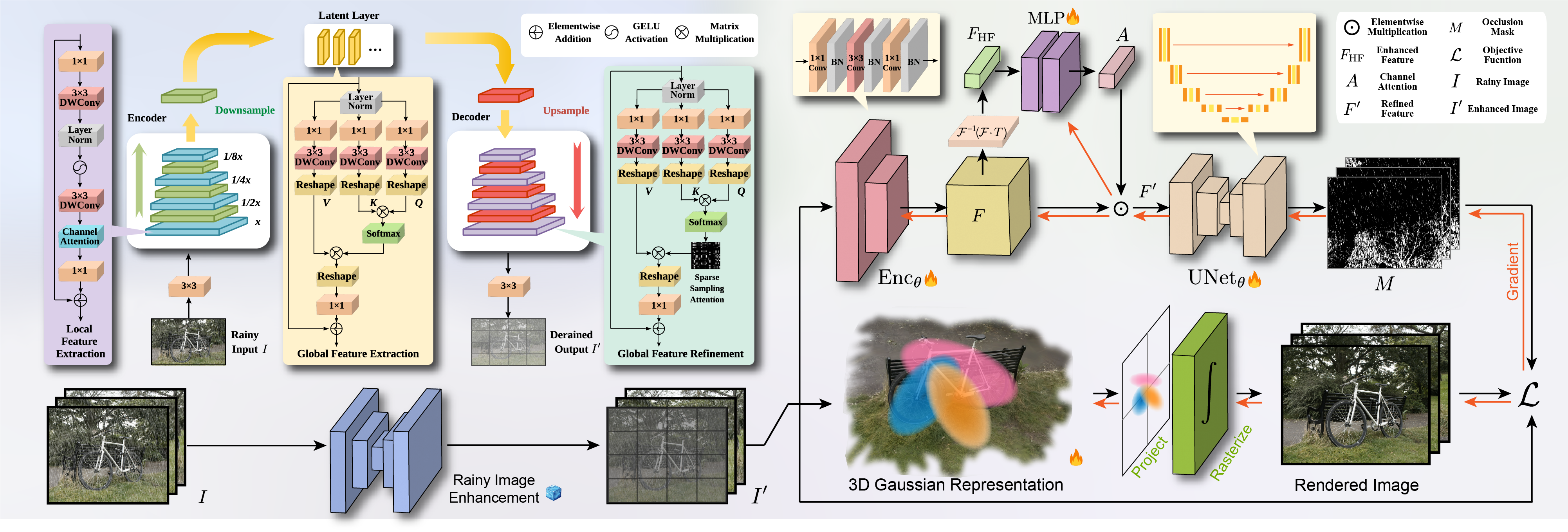}
\caption{The illustration of the DeRainGS pipeline. The left part displays the rainy image enhancement procedure and its network structure, which is pretrained before being applied to a scene. The right part demonstrates the reconstruction by 3DGS, utilizing learned occlusion masks to handle rain-induced artifacts.}
\label{fig:pipeline}
\end{figure*}

\subsection{Real-world Data Collection}

We collect videos of real-world rainy scenes using a SONY-A7R3 camera in Full HD format. The scene images are extracted from the video at a rate of one frame per second. The footage is captured across six different scenes in Tokyo, with most scenes specifically filmed in the evening to enhance the contrast of rain streaks through ambient light. To ensure privacy protection, all pedestrians and any identifiable personal elements are manually erased using Photoshop software.

\section{Proposed Method}

\subsection{Preliminaries: 3D Gaussian Splatting (3DGS)}

3DGS \cite{kerbl20233d} is a breakthrough method for reconstructing and rendering 3D scenes from an image collection. It employs a set of 3D Gaussian SHs ${G_i}$ to explicitly represent the scene. Each Gaussian $G_i$ is characterized by its mean position $\mu_i \in \mathbb{R}^3$, a 3D covariance matrix $\sigma_i \in \mathbb{R}^{3\times3}$, an opacity $\alpha_i$, and view-dependent color coefficients $c_i$. The mean positions $\mu_i$ of the Gaussian points are initialized using point clouds derived from Structure from Motion (SfM) \cite{schonberger2016structure}. The covariance matrix $\Sigma_i$ is decomposed into a scaling matrix $S$ and a rotation matrix $R$ to ensure positive semi-definiteness:
\begin{equation}
\Sigma_i = RS S^T R^T~.
\end{equation}
During rendering, these 3D Gaussians are projected onto the image plane. Given a world-to-camera transformation $W$ and the Jacobian $J$ of an affine approximation of perspective projection, the 2D covariance matrix $\Sigma'_i$ is computed as:
\begin{equation}
\Sigma'_i = (JW\Sigma_i W^T J^T)~.
\end{equation}
The color of each pixel is then determined through alpha compositing, traversing the Gaussians from front to back
\begin{equation}
C = \sum_i c_i(\mathbf{r}) \alpha_i' \prod_{j=1}^{i-1} (1 - \alpha_j')~.
\end{equation}
In this context, $c_i(\mathbf{r})$ is the view-dependent color computed using SHs, and $\alpha_i'$ is the blending weight equal to the product of the Gaussian's opacity and its 2D projection.


\subsection{Rainy Image Enhancement}
\label{sec:derain_nn}
We first perform rainy image enhancement as a preprocessing step to tackle rain effects, providing an enhanced image to guide subsequent scene reconstruction. To enhance the robustness of the deraining model across different scenes, we model complex rain distribution by combining local and non-local information. In the deraining network, we utilize a 5-level encoder-decoder architecture. The overall framework is presented on the left of \Cref{fig:pipeline}. Assuming \(I_k\) is the original image, it undergoes local feature extraction, global feature extraction, and global feature refinement stages, resulting in \(I_k'\) as the enhanced image.

For local feature extraction, it contains $\{4,6,7,8\}$ convolutional blocks in the network encoder. Specifically, we begin by doubling the channel dimension, and then use the DWConvLN-GELU-DWConv design to capture rain information. In the bottleneck layer of the encoder-decoder, we stack 8 vanilla Transformer modules~\cite{xiao2022image} for global deep feature extraction. For the network decoder, we adopt $\{3,6,7,8\}$ sparse Transformer modules~\cite{chen2023sparse} to reconstruct the final outputs. Here, sparse Transformer modules are employed to eliminate irrelevant feature interactions among tokens to generate high-quality outputs, while vanilla Transformer modules are employed to ensure necessary information flows through the whole network. In summary, our goal is to effectively integrate their complementary features by CNN and Transformer to better achieve comprehensive rain distribution prediction.

The end-to-end rainy image enhancement network is trained on the rain streak dataset 4K-Rain13k~\cite{chen2024towards} and raindrop dataset UAV-Rain1k~\cite{chang2024uav}, and the model is frozen during the reconstruction process.

\subsection{Scene Reconstruction}
\label{sec:scene_recon}

\begin{figure}[!ht]
\centering
\includegraphics[width=0.45\textwidth]{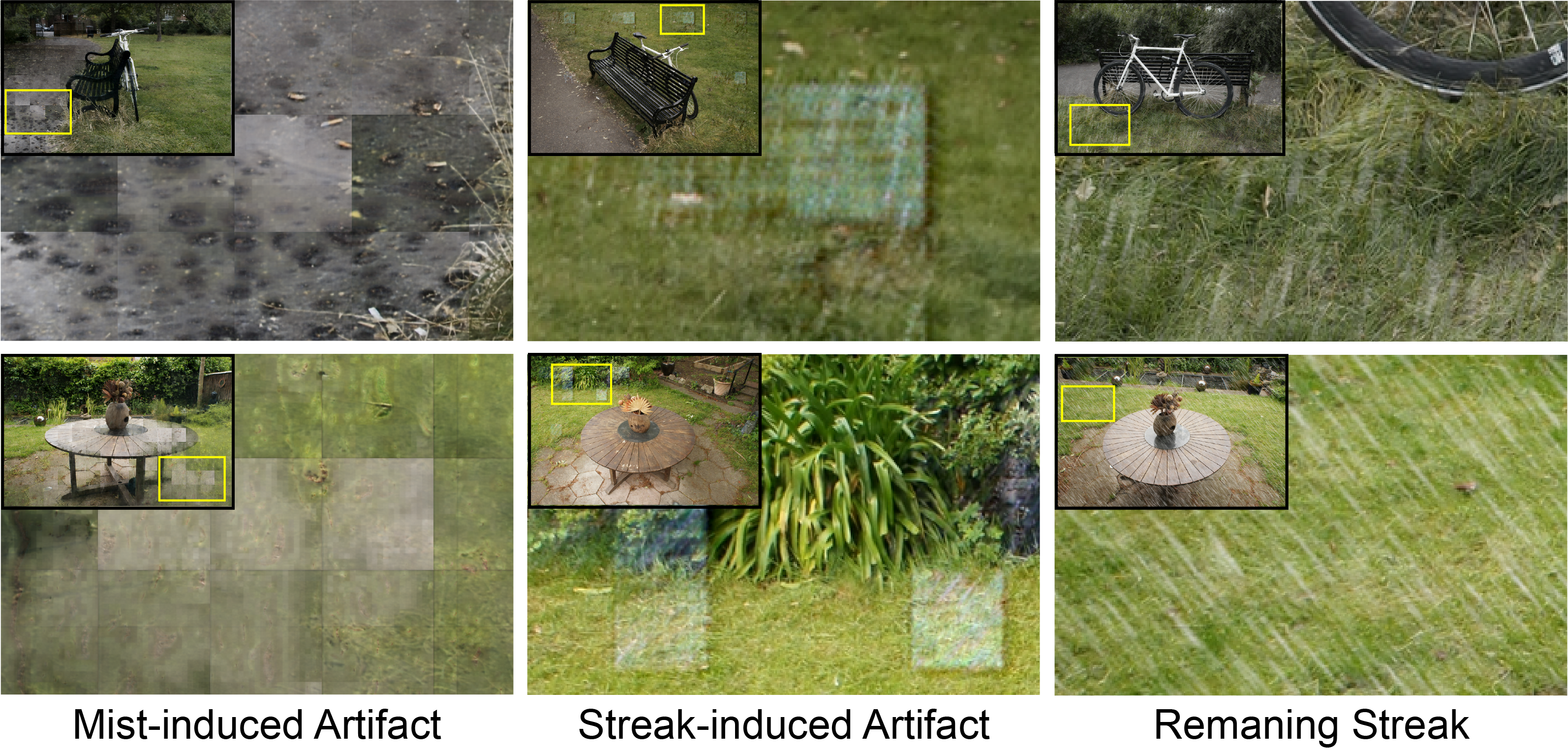}
\caption{Different types of artifacts, produced either by the image enhancement process or the persisting rain effect.}
\label{fig:derainer-artifact}
\end{figure}

The varied shapes and distortions caused by rain pose substantial challenges for a derainer, making the complete removal of rain effects extremely difficult. Moreover, beyond the residual effects of rain, the end-to-end enhancement process may also introduce additional artifacts as a result of over-correction for distortion. As shown in \Cref{fig:derainer-artifact}, these artifacts, which often appear as high-frequency patterns, can adversely impact the quality of the reconstructed scene.

To address the intricate problem of high-frequency artifacts, we propose an unsupervised learning approach to predict masks of artifacts. As depicted in the right of \Cref{fig:pipeline}, our method strategically leverages spectral pooling within a channel attention module, which is designed to enhance sensitivity to high-frequency details that could potentially manifest as artifacts. Thereafter, refined features are processed by a U-Net model to generate masks. Unlike typical employment of U-Net within in-the-wild approaches which either segment by categories \cite{xu2024wild, yang2023cross} or learn the static content of the scene \cite{wang2024we}, our network is optimized for the direct identification of rain artifacts. This design leverages the fact that raindrops or mist on the lens typically maintain a fixed position relative to viewpoints, and rain streaks usually display a consistent pattern across one scene, despite their varied shapes, making them feasible to capture.

\subsubsection{Frequency-based Feature Channel Attention}

The initial step in the pipeline involves a CNN encoder $ \text{Enc}_{\theta}$ that processes the enhanced image $I_k'$ and produces a feature map $F_k \in \mathbb{R}^{C \times H' \times W'}$. The encoder captures a broad array of features across different channels. Following feature extraction, we utilize spectral pooling to manipulate these features to accentuate high-frequency components that are crucial for detecting artifacts. This process involves transforming the $F_k$ into the frequency domain using the Fourier Transform:
\begin{equation}
\tilde{F}_k = \mathcal{F}(F_k) = \sum_{x=0}^{H'-1} \sum_{y=0}^{W'-1} F_k[x, y] \cdot e^{-2\pi i \left(\frac{ux}{H'} + \frac{vy}{W'}\right)}
\end{equation}

\noindent where $\mathcal{F}$ denotes the Fourier Transform applied to each channel of $F_k$ and produces frequency components $\tilde{F}_k \in \mathbb{R}^{C \times H' \times W'}$. In the frequency domain, a high-pass filtering operation is employed to isolate and enhance high-frequency features that are potentially indicative of artifacts:
\begin{equation}
\tilde{F}_{\text{HF}} = \tilde{F}_k \cdot T
\end{equation}

\noindent $T$ represents the high-pass filter mask. Subsequently, we utilize the Inverse Fourier Transform to convert these emphasized frequency-domain features back to the spatial domain:
\begin{equation}
F_{\text{HF}} = \mathcal{F}^{-1}(\tilde{F}_{\text{HF}}) = \frac{1}{H'W'} \sum_{u=0}^{H'-1} \sum_{v=0}^{W'-1} \tilde{F}_{\text{HF}}[u, v] \cdot e^{2\pi i \left(\frac{ux}{H'} + \frac{vy}{W'}\right)}
\end{equation}

\noindent The resulting feature map $F_{\text{HF}} \in \mathbb{R}^{C \times H' \times W'}$ predominantly contains high-frequency information that corresponds to potential artifacts in the image.

A channel attention mechanism is then employed to refine the focus on these high-frequency details. An MLP processes the high-frequency emphasized features $F_{\text{HF}}$ to compute channel-wise attention scores:
\begin{equation}
A_k = \text{Sigmoid}(\text{MLP}_{\theta}(F_{\text{HF}}))
\end{equation}

\noindent The attention scores $A_k \in \mathbb{R}^{C \times 1 \times 1}$ serve to modulate the significance of each channel in the original feature map \( F_k \) based on their relevance to high-frequency features.

\subsubsection{Mask Prediction}

We apply the attention scores $A_k$ back to the original feature map to enhance channels identified as significant, resulting in a refined feature map:
\begin{equation}
F'_k = A_k \odot F_k
\end{equation}

\noindent The refined feature map $F'_k \in \mathbb{R}^{C \times H' \times W'}$, is then fed into a U-Net architecture to predict a mask $M_k'$:
\begin{equation}
M_k' = \text{UNet}_{\theta}(F'_k)
\end{equation}

\noindent Subsequently, the mask $M_k'$ is resized to $M_k \in \mathbb{R}^{1 \times H \times W}$, aligning with the dimensions of the original image $I_k$. The mask $M_k$ aims to highlight the locations and extents of high-frequency artifacts, which is crucial for excluding erroneous pixels that could degrade the quality of reconstruction through optimizing photometric residuals.

\begin{figure*}[ht]
\centering
\includegraphics[width=\textwidth]{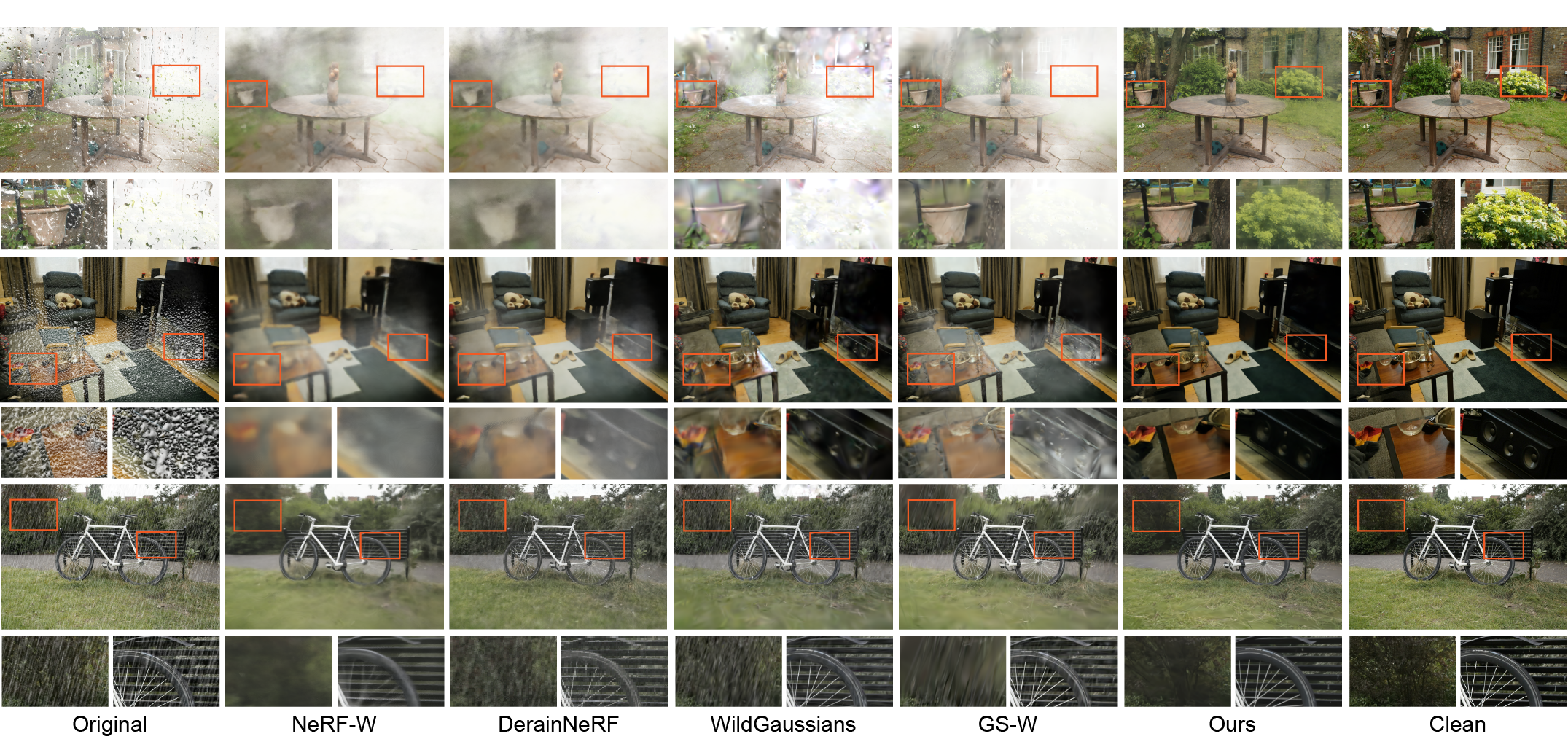}
\caption{The visualization compares the rendering outcomes of DeRainGS with baseline methods on selected scenes featuring raindrops (top two rows) or rain streaks (bottom row) of our HydroViews dataset.}
\label{fig:ras-figure}
\end{figure*}

\subsubsection{Optimization}

DeRainGS is trained in an end-to-end manner. The rainy image enhancement network is frozen during the training. We optimize all parameters of Gaussians $\mathcal{G}$, the feature extractor $\text{Enc}_{\theta}$, the channel attention layer $\text{MLP}_{\theta}$, and the mask predictor $\text{UNet}_{\theta}$. The training objective for the 3DGS scene representation is defined as:
\begin{align}
\mathcal{L}_C &= (1 - \lambda_{\rm SSIM})(1 - M) \odot \| \hat{C} - C \| \\
&+ \lambda_{\rm SSIM} (1 - M) \odot \mathcal{L}_{\rm SSIM}(\hat{C},C)
\end{align}

\noindent where the mask $M$ is integrated within the photometric loss to specifically exclude pixels affected by rain. The SSIM loss is crucial for improving the structural accuracy of the rendered scenes, which are broadly affected by the rain effect. The final objective function can be formalized as:
\begin{equation}
\mathcal{L} = \mathcal{L}_C + \lambda_{\rm reg} \mathcal{L}_{\rm reg}
\end{equation}

\noindent where $\mathcal{L}_{\rm reg} = \sum_{i}^{H} \sum_{j}^{W}{M_{ij}^2}$ is the mask regularization loss to prevent the predicted masks from excessively masking everything. $\lambda_{\rm SSIM}$ and $\lambda_{\rm reg}$ are weighting hyperparameters.

\begin{table*}[!t]
\centering
\caption{The quantitative evaluation of our method and baseline approaches on the HydroViews dataset. Results are averaged over three different types of raindrops or rain streaks within each scene. An underline indicates one or more failures occurred in the reconstruction process among three trials of different rain effects for a scene.}
\small 
\begin{tabular}{@{}c@{\hspace{2pt}}c@{\hspace{3pt}}|l@{\hspace{12pt}}|c@{\hspace{6pt}}c@{\hspace{6pt}}c@{\hspace{8pt}}c@{\hspace{6pt}}c@{\hspace{6pt}}c@{\hspace{8pt}}c@{\hspace{6pt}}c@{\hspace{6pt}}c@{\hspace{8pt}}c@{\hspace{6pt}}c@{\hspace{6pt}}c@{\hspace{8pt}}c@{\hspace{6pt}}c@{\hspace{6pt}}c@{}}
\toprule
&&& \multicolumn{3}{c}{NeRF-W} & \multicolumn{3}{c}{DerainNeRF} & \multicolumn{3}{c}{WildGaussians} & \multicolumn{3}{c}{GS-W} & \multicolumn{3}{c}{\textbf{Ours}} \\ 
\cmidrule(lr){4-6} \cmidrule(lr){7-9} \cmidrule(lr){10-12} \cmidrule(lr){13-15} \cmidrule(lr){16-18}
 && Scene & PSNR & SSIM & LPIPS & PSNR & SSIM & LPIPS & PSNR & SSIM & LPIPS & PSNR & SSIM & LPIPS & PSNR & SSIM & LPIPS \\ 
 \midrule
 && Bicycle & 12.88 & 0.317 & 0.622 & 12.72 & 0.367 & 0.595 & 12.25 & 0.269 & 0.664 & 13.27 & 0.416 & 0.507 & 22.18 & 0.687 & 0.307 \\
 && Bonsai & 17.16 & 0.554 & 0.458 & 17.66 & 0.606 & 0.422 & 18.88 & 0.713 & 0.27 & 18.78 & 0.749 & 0.238 & 27.41 & 0.877 & 0.173 \\
 & \multirow{3}{*}{\rotatebox[origin=c]{90}{Raindrop}}
 & Counter & 19.23 & 0.617 & 0.414 & 20.24 & 0.701 & 0.358 & 22.22 & 0.741 & 0.281 & 22.51 & 0.769 & 0.239 & 28.83 & 0.894 & 0.137 \\
 && Garden & 8.43 & 0.282 & 0.667 & 8.52 & 0.354 & 0.629 & 10.73 & 0.396 & 0.510 & 10.11 & 0.417 & 0.468 & 22.89 & 0.692 & 0.278 \\
 && Kitchen & 16.54 & 0.522 & 0.421 & 17.88 & 0.598 & 0.409 & 17.02 & 0.729 & 0.208 & \underline{16.73} & \underline{0.751} & \underline{0.204} & 26.27 & 0.854 & 0.141 \\
 && Room & 18.97 & 0.640 & 0.395 & 15.53 & 0.645 & 0.355 & 23.21 & 0.754 & 0.324 & 22.37 & 0.760 & 0.278 & 31.28 & 0.899 & 0.180 \\
 && Stump & 18.54 & 0.368 & 0.617 & 18.68 & 0.408 & 0.596 & \underline{18.31} & \underline{0.468} & \underline{0.501} & 20.46 & 0.520 & 0.487 & 28.69 & 0.841 & 0.200 \\
 \midrule
 && Bicycle & 17.80 & 0.386 & 0.575 & 13.24 & 0.286 & 0.665 & 16.48 & 0.417 & 0.504 & 18.20 & 0.459 & 0.489 & 26.52 & 0.834 & 0.220 \\
 & \multirow{5}{*}{\rotatebox[origin=c]{90}{Rain streak}}
 & Garden & 19.78 & 0.474 & 0.474 & 15.25 & 0.393 & 0.566 & 18.55 & 0.604 & 0.329 & 19.89 & 0.621 & 0.338 & 25.42 & 0.705 & 0.216 \\
 && Kitchen & 19.46 & 0.622 & 0.373 & 17.35 & 0.602 & 0.364 & 20.12 & 0.750 & 0.230 & \underline{20.39} & \underline{0.783} & \underline{0.228} & 27.40 & 0.800 & 0.164 \\
 && Stump & 17.60 & 0.372 & 0.586 & 14.19 & 0.318 & 0.554 & 16.60 & 0.411 & 0.564 & 17.87 & 0.423 & 0.530 & 27.11 & 0.773 & 0.234 \\
 && Horse & 17.76 & 0.756 & 0.375 & 14.97 & 0.578 & 0.516 & 17.85 & 0.788 & 0.337 & 19.32 & 0.789 & 0.328 & 27.20 & 0.889 & 0.194 \\
 && Train & 17.42 & 0.607 & 0.456 & 14.39 & 0.512 & 0.542 & 17.20 & 0.618 & 0.449 & 19.18 & 0.643 & 0.414 & 24.74 & 0.818 & 0.264 \\
 && Truck & 19.34 & 0.700 & 0.442 & 16.36 & 0.609 & 0.503 & 19.89 & 0.747 & 0.360 & 20.05 & 0.727 & 0.378 & 26.40 & 0.895 & 0.216 \\
\bottomrule
\label{tab:hydroviews}
\end{tabular}
\end{table*}

\section{Experiments}


\subsubsection{Datasets}
We evaluate our method on the waterdrop scenes from DerainNeRF \cite{li2024derainnerf} and our HydroViews dataset. In our HydroViews dataset, each synthesized scene features three distinct patterns of raindrops and rain streaks. We evaluate all methods across three patterns and report the averaged results.

\subsubsection{Metrics}
To assess the quality of the reconstruction, we adhere to common metrics including PSNR, SSIM, and LPIPS by comparing the rendered scenes with the clean images.

\subsubsection{Baselines}
We compare our method with deraining and in-the-wild reconstruction methods, including NeRF-W \cite{martin2021nerf}, DerainNeRF \cite{li2024derainnerf}, GS-W \cite{zhang2024gaussian} and WildGaussians \cite{kulhanek2024wildgaussians}. In the case of DerainNeRF \cite{li2024derainnerf}, which employs an AttGAN \cite{qian2018attentive} derainer, the weights of the original network are not publicly available. To ensure a fair comparison, we retrained the AttGAN network on the same datasets, 4K-Rain13k \cite{chen2024towards} and UAV-Rain1k \cite{chang2024uav}, that were used to train our rainy image enhancement network.

\subsubsection{Implementation Details}
We implement DeRainGS in PyTorch using Adam optimizer with learning rates of 1.6e-4, 5e-4, and 2.5e-3 for $\mathcal{G}$'s means, scaling, and SH features, and 1e-3 for tuning occlusion masking modules. We minimize $\mathcal{L}$ over 70,000 iterations. All experiments were conducted on a single NVIDIA A100-80GB GPU. Additional implementation details are provided in the supplementary material.


\begin{table*}[t]
\centering
\caption{The quantitative evaluation of our method and baseline approaches on the DerainNeRF dataset \cite{li2024derainnerf}.}
\small 
\begin{tabular}{@{}c@{\hspace{2pt}}c@{\hspace{3pt}}|l@{\hspace{3pt}}|c@{\hspace{6pt}}c@{\hspace{6pt}}c@{\hspace{9pt}}c@{\hspace{6pt}}c@{\hspace{6pt}}c@{\hspace{9pt}}c@{\hspace{6pt}}c@{\hspace{6pt}}c@{\hspace{9pt}}c@{\hspace{6pt}}c@{\hspace{6pt}}c@{\hspace{9pt}}c@{\hspace{6pt}}c@{\hspace{6pt}}c@{}}
\toprule
&&& \multicolumn{3}{c}{NeRF-W} & \multicolumn{3}{c}{DerainNeRF} & \multicolumn{3}{c}{WildGaussians} & \multicolumn{3}{c}{GS-W} & \multicolumn{3}{c}{\textbf{Ours}} \\ 
\cmidrule(lr){4-6} \cmidrule(lr){7-9} \cmidrule(lr){10-12} \cmidrule(lr){13-15} \cmidrule(lr){16-18}
 & & Scene & PSNR & SSIM & LPIPS & PSNR & SSIM & LPIPS & PSNR & SSIM & LPIPS & PSNR & SSIM & LPIPS & PSNR & SSIM & LPIPS \\ 
\midrule
 & \multirow{4}{*}{\rotatebox[origin=c]{90}{Waterdrop}}
 & Church & 29.76 & 0.934 & 0.071 & 30.10 & 0.968 & 0.043 & 32.88 & 0.975 & 0.070 & 31.68 & 0.970 & 0.068 & 33.23 & 0.975 & 0.063 \\
 && Factory & 29.03 & 0.896 & 0.078 & 30.85 & 0.923 & 0.056 & 31.70 & 0.965 & 0.081 & 31.25 & 0.958 & 0.076 & 32.67 & 0.971 & 0.074 \\
 && Tanabata & 25.58 & 0.873 & 0.121 & 26.04 & 0.887 & 0.108 & 33.58 & 0.975 & 0.077 & 33.16 & 0.973 & 0.071 & 34.62 & 0.980 & 0.068 \\
 && Doraemon & 27.83 & 0.894 & 0.117 & 28.74 & 0.915 & 0.102 & 33.11 & 0.957 & 0.085 & 32.46 & 0.951 & 0.092 & 33.94 & 0.967 & 0.076 \\
\bottomrule
\label{tab:derainnerf}
\end{tabular}
\end{table*}

\subsection{Raindrops and Rain Streaks}

\Cref{tab:hydroviews} presents the quantitative comparison of our DeRainGS model with baseline approaches on our HydroViews dataset. In challenging raindrop scenarios, such as scenes containing large occlusions like bicycle and garden, DeRainGS benefits from the effective rainy image enhancement strategy and consistently delivers high-quality reconstructions, surpassing baseline methods that struggle with mist on the lens. For scenes affected by rain streaks, DeRainGS delivers even higher performance in scene recovery. In contrast, DerainNeRF \cite{li2024derainnerf}, which lacks occlusion handling, suffers from inconsistent geometry and performs even worse than NeRF-W \cite{martin2021nerf}. It is important to note that the exceptional performance of DeRainGS is not primarily derived from the high-fidelity rendering capabilities of 3DGS, as the performance of the 3DGS baselines is on par with that of the NeRF approaches. This indicates that the enhancements and masking modules specific to our method are crucial in achieving superior results. \Cref{fig:ras-figure} visualizes selected reconstruction outcomes, demonstrating that our method achieves rendering quality close to clean scenes. In contrast, the baseline methods struggle with distortion and occlusion induced by rainstreak and mist.

Furthermore, \Cref{tab:derainnerf} presents the rendering evaluation on the dataset released by DerainNeRF \cite{li2024derainnerf}. DeRainGS continues to exhibit state-of-the-art performance compared to the NeRF and 3DGS-based approaches. However, since the dataset predominantly features lightweight waterdrops, the differences in performance between each method are marginal. This underscores the need to construct more realistic datasets for benchmarking.

\subsection{Real-world Rainy Environments}

\begin{figure}[H]
\centering
\includegraphics[width=0.45\textwidth]{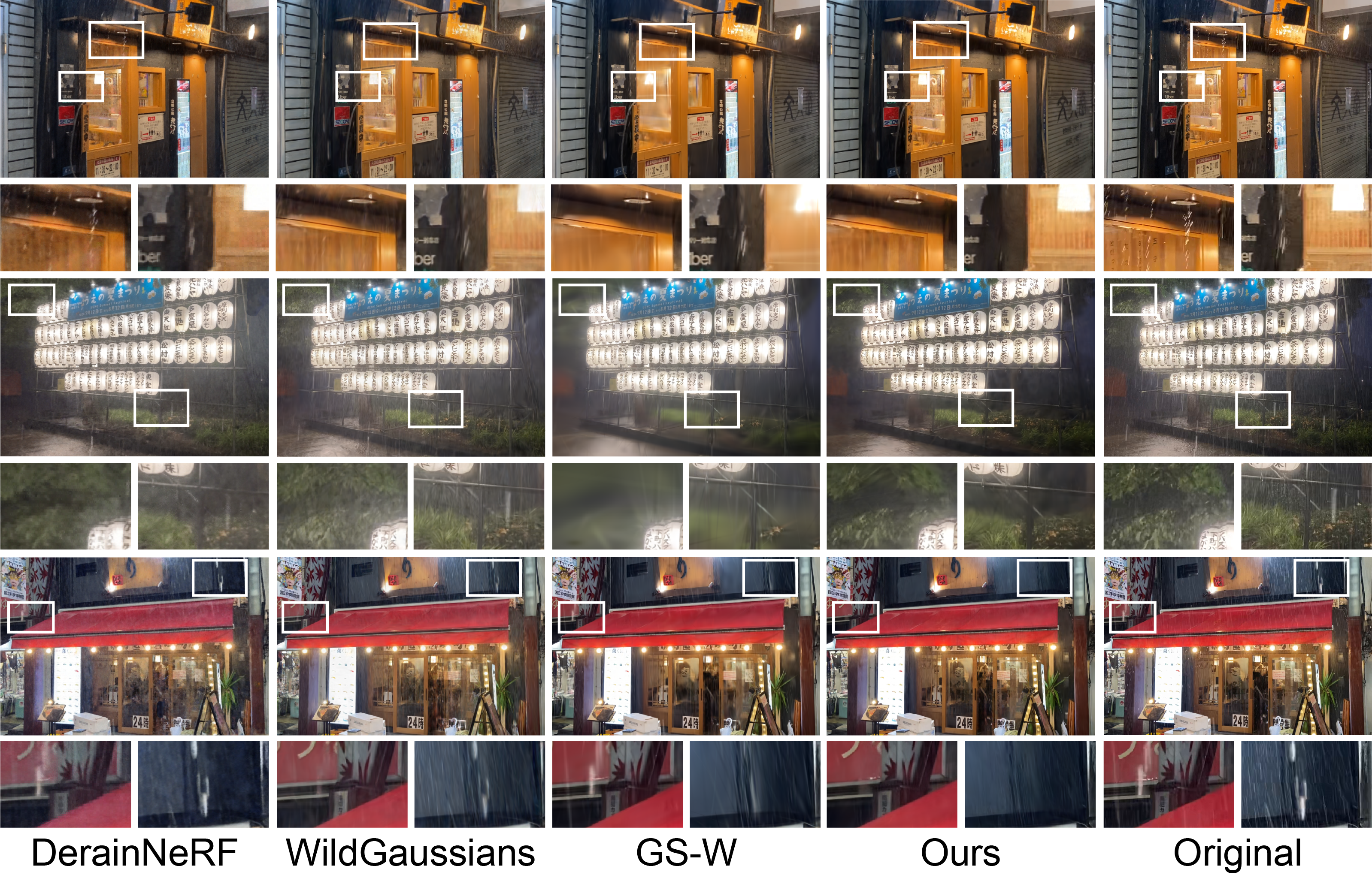}
\caption{The visualization compares the rendering quality of DeRainGS and baselines on real-world scenes.}
\label{fig:realworld-figure}
\end{figure}

To demonstrate the generalization of DeRainGS, \Cref{fig:realworld-figure} compares our method with baseline approaches on selected real-world scenes from our HydroViews dataset. These real-world scenes present significant challenges posed by irregularly shaped rain streaks. DeRainGS effectively mitigates the rain effect, surpassing both NeRF and 3DGS-based methods, which struggle with accurate scene reconstruction and tackling complex rain effects.

\subsection{Ablation Study}

\begin{figure}[t]
\centering
\includegraphics[width=0.45\textwidth]{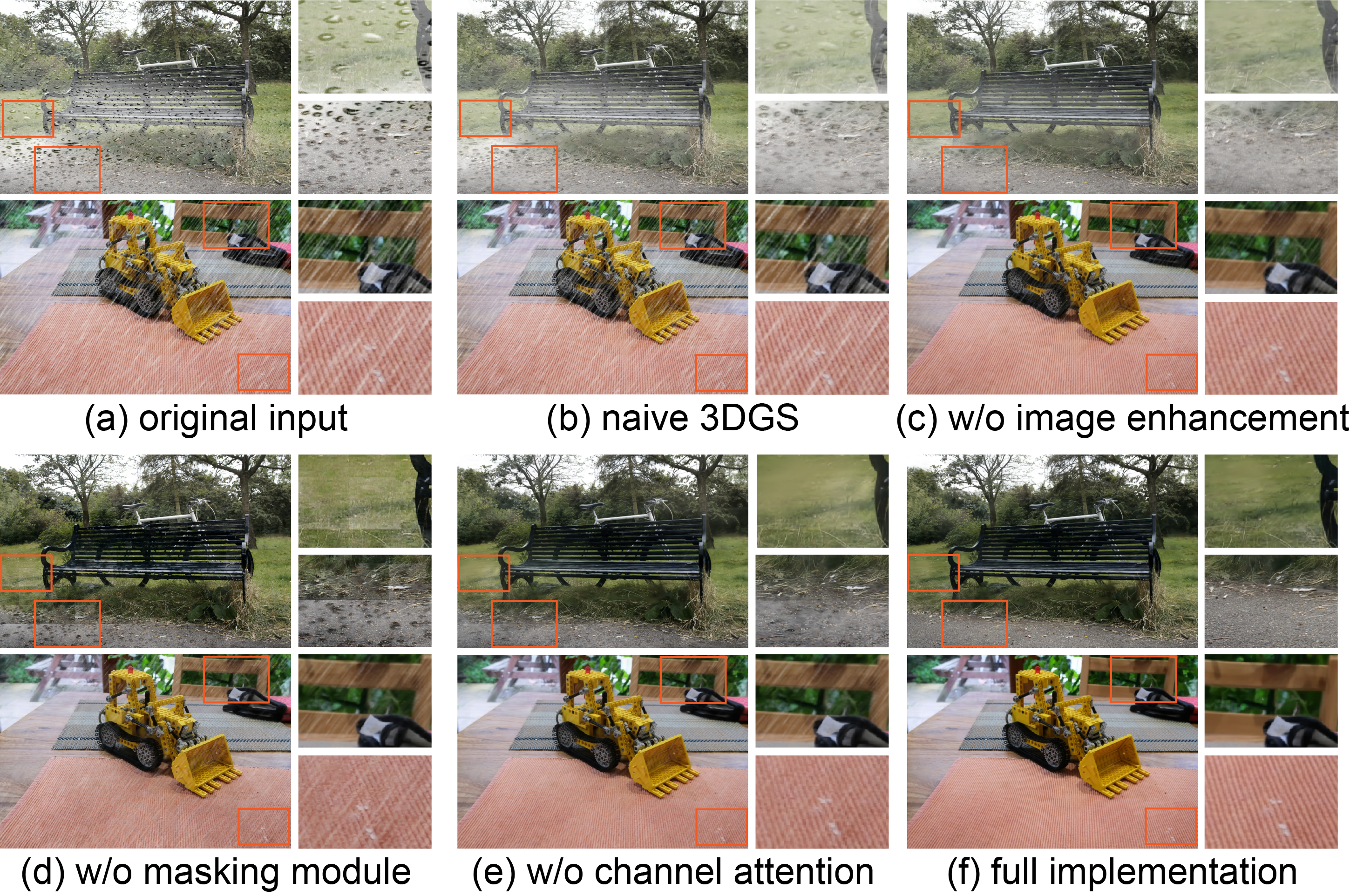}
\caption{Visualization of rendering quality for ablations.}
\label{fig:ablation-figure}
\end{figure}

We conduct ablation studies on both the raindrop and rain streak scenes for each component of DeRainGS and present the average results for quantitative evaluation. As indicated in \Cref{tab:ablation}, we find that utilizing either the enhancement process as in (d) or the occlusion masking strategy as in (c) can individually improve the reconstruction quality to a certain degree. 
Moving to a full implementation as in (f) can thereby significantly enhance performance when compared with vanilla 3DGS. When comparing (c) and (f), we notice that the frequency-based channel attention module also plays a key role in effectively predicting rain occlusions. To straightforwardly show each component's contribution, \Cref{fig:ablation-figure} illustrates the rendering outcomes of configurations (b) to (f). We observe that the enhancement process effectively mitigates mist caused by raindrops in (d), while the masking strategy with the channel attention module can adeptly handle high-frequency rain streaks, as shown in (c).

\begin{table}[tb]
\centering
\caption{Ablation study of each component of DeRainGS.}
\label{tab:sample_table}
\small
\setlength{\tabcolsep}{2pt}
\begin{tabular}{lcccccc}
\toprule
 & \multicolumn{4}{c}{DeRainGS's Components} & \multicolumn{2}{c}{Metrics} \\
\cmidrule(lr){2-5} \cmidrule(lr){6-7}
 & 3DGS & Image Enh. & Mask Mod. & Chan. Attn. & PSNR / SSIM & \\
\midrule
(b) & \checkmark & & & & 17.56 / 0.659 & \\
(c) & \checkmark & & \checkmark & \checkmark & 19.89 / 0.726 & \\
(d) & \checkmark & \checkmark & & & 20.42 / 0.721 & \\
(e) & \checkmark & \checkmark & \checkmark & & 23.25 / 0.770 & \\
(f) & \checkmark & \checkmark & \checkmark & \checkmark & 24.67 / 0.782 & \\
\bottomrule
\label{tab:ablation}
\end{tabular}
\end{table}

\section{Conclusion}
In this study, we introduce the new task of 3DRRE, aiming to tackle the complex challenges of 3D scene reconstruction under adverse rainy conditions. To support this initiative, we have developed the HydroViews dataset—a comprehensive benchmark that includes both synthesized and real-world scenes affected by various rain effect. Building upon this foundation, we propose DeRainGS, the first 3DGS-based method tailored specifically for reconstruction in rainy environments. DeRainGS significantly outperforms existing occlusion-free baselines under various raindrops and rain streak conditions. Together, these contributions enhance the capability to reconstruct high-quality 3D scenes in rain-impacted environments and pave the way for further research in weather-adaptive 3D reconstruction techniques.

\bibliography{aaai25}

\end{document}